\pdfoutput=1
\documentclass[11pt]{article}

\usepackage[final]{acl}

\usepackage{times}
\usepackage{latexsym}

\usepackage[T1]{fontenc}
\usepackage[utf8]{inputenc}

\usepackage{microtype}

\usepackage{inconsolata}
\usepackage{amsmath}
\usepackage{amssymb}
\usepackage{mathtools}
\usepackage{amsthm}
\usepackage[english]{babel}
\usepackage{booktabs}
\usepackage{xcolor} 
\title{Causal-Guided Active Learning for Debiasing Large Language Models}

\author{Li Du$^2$\thanks{This is an informal Edition.}, Zhouhao Sun${^1}^*$\thanks{Equal contribution. Listing order is random.}, Xiao Ding$^1$\thanks{Corresponding Author}, Yixuan Ma$^1$, { \bf Yang Zhao$^1$, Kaitao Qiu$^3$, Ting Liu$^1$, Bing Qin$^1$} \\
$^1$Research Center for Social Computing and Information Retrieval\\
Harbin Institute of Technology, China\\
$^2$Beijing Academy of Artificial Intelligence, Beijing, China \\
$^3$ Harbin Institute of Technology, China \\
\{zhsun, xding, yxma, yzhao, tliu, bqin\}@ir.hit.edu.cn\\
duli@baai.ac.cn\\ astro073@outlook.com\\
}
\newenvironment{shrinkeq}[1]
{ \bgroup
\addtolength\abovedisplayskip{#1}
\addtolength\belowdisplayskip{#1}}
{\egroup\ignorespacesafterend}

\begin{document}
\maketitle
\begin{abstract}
Although achieving promising performance, recent analyses show that current generative large language models (LLMs) may still capture dataset biases and utilize them for generation, leading to poor generalizability and harmfulness of LLMs. However, due to the diversity of dataset biases and the over-optimization problem, previous prior-knowledge-based debiasing methods and fine-tuning-based debiasing methods may not be suitable for current LLMs.
To address this issue, we explore combining active learning with the causal mechanisms and propose a casual-guided active learning (CAL) framework, which utilizes LLMs itself to automatically and autonomously identify informative biased samples and induce the bias patterns. Then a cost-effective and efficient in-context learning based method is employed to prevent LLMs from utilizing dataset biases during generation.
Experimental results show that CAL can effectively recognize typical biased instances and induce various bias patterns for debiasing LLMs.
\end{abstract}

\section{Introduction}
% % \vspace{-0.2cm}
Large language models (LLMs) are growing to be the foundation of Natural Language Processing. Through the generative pretraining process upon a large-scale corpus, the LLMs have demonstrated impressive performance in understanding the language and conducting complex reasoning tasks \cite{achiam2023gpt}, %Then by fine-tuning LLMs on instruction sets, LLMs are further empowered with the ability to follow human instructions \cite{chung2022scaling,ouyang2022training}, reforming the interaction paradigm between humans and models, 
demonstrating immense potential in real-world applications.

However, the generative pretraining process is a double-edged sword, as it would also inevitably incur \textbf{dataset bias} into the LLMs such as position bias and stereotype bias \cite{schick2021self,navigli2023biases,zheng2023judging,shaikh-etal-2023-second}. This is because, the LLMs only \emph{passively} learn to model the \emph{correlation} between contexts in the pretraining corpus, and the pretraining corpus is biased as it reflects the inherent preference or prejudice of human beings.  
%Dataset bias refers to the unwanted correlation between certain patterns among the training corpus. 
For example, the existence of position bias is due to the subconscious human belief that the first option is better, leading to a higher frequency of the first option in corpora, and LLMs trained to model the corpus distribution would also capture such biased correlation. 
%Similarly, the presence of stereotype bias such as gender bias stems from the societal belief that one gender is superior to or more suitable for certain roles than another, which is also reflected in the corpus through uneven representation or stereotypical portrayal. However, the suitability or superiority of an individual for certain roles or tasks is entirely independent of their gender.
%if the person's name of a context is usually associated with males, there is a significantly higher likelihood that subsequent texts will attribute strong science skills to them, in contrast to individuals with names typically associated with females. Since the generative LLMs learns to generate the subsequent based on context, such connection between the 
%Consequently, generative LLMs which learns to generate the subsequent text may incur or even amplify this bias from the data. 
%However, these achievements stem from training on extensive text corpora \cite{zhao2023survey}, which inherently possess biases that are correlated with the inference target but do not represent a causal relationship \cite{navigli2023biases}. Consequently, LLMs may inadvertently internalize these biases during training.  
Such biases would lead to \emph{poor generalizability} and \emph{harmfulness} of LLMs \cite{navigli2023biases,huang2023trustgpt}. 
For instance, when an LLM is asked to evaluate which option is better, the LLM may utilize position bias and tend to choose the first option. However, which option is better is completely unrelated to its position. Therefore, when the second option is generally better in some datasets, the performance of the LLM will significantly decline. While biases such as stereotyping bias would make LLMs generate harmful content such as women are less capable in STEM fields, which in turn reinforces harmful stereotypes. 

These problems highlight the necessity of debiasing LLMs.
The key issue to debias LLMs lies in how to recognize the dataset biases and prevent it from utilizing biases during inference. To this end, prevalent methods rely on researchers' prior knowledge to artificially recognize the potential dataset biases, and then eliminate such biases through aligning or prompt-based regularization \cite{schick2021self,oba2023contextual,liu2023trustworthy}. However, due to the diversity and complexity of dataset biases \cite{poliak2018hypothesis,schuster2019towards,schick2021self}, it's impractical to identify them one by one manually. A vast amount of biases remains unrecognized in different tasks \cite{nie2020adversarial} and new biases are continually being discovered. 
%For example, \citet{zhang2023ethical} recently recognized a subtle ``Grandma Bias'', i.e., GPT-4 might learn the dataset bias that grandmothers tend to spoil their grandsons. Consequently, if GPT-4 is prompted to simulate a grandmother's behavior and coax her grandson by using certain illegal information, GPT-4 might violate the restriction of obeying law content and generate illegal information. 

%another line of work proposes to automatically identify the dataset biases by extracting features that can be empirically deemed to characterize the distribution of dataset biases through training certain ``biased models'' \cite{utama2020towards,sanh2020learning,du2023towards}. Then in a following fine-tuning process, the model can be regularized in the feature space to prevent it from capturing the biased information. 
Hence, there is an urgent need for methods to automatically identify biases of generative LLMs. However, previous automatic debiasing methods are mainly designed for discriminative models and are hard to adapt to generative LLMs. Moreover, these methods generally rely on a fine-tuning-based process on certain dataset(s) to regularize the model. The finetuning-based debiasing process would lead to over-optimization and undermine the generalizability of LLMs on other tasks \cite{aribandi2021ext5,liu2023mftcoder}.
%, and the fine-tuning process of LLMs would be rather costly.

To address these issues, considering the powerful pattern recognition and inductive ability of LLMs, we explore combining \emph{active} learning with the \emph{causal} mechanisms and propose a \textbf{C}asual-guided \textbf{A}ctive \textbf{L}earning (CAL) framework, which utilizes LLMs themselves to automatically and autonomously identify biased samples and induce the bias patterns. 
Active learning aims at selecting the most informative instances, and then querying external information source(s) to label these data points. In the debiasing scenario, 
CAL identifies the biased instances by finding instances where the LLMs fail to model \emph{causal invariant} semantic relationship among context, then selects the most informative biased instances by finding the instances on which dataset biases have the most influence on the generation of LLMs. The causal invariance can be employed to disentangle the semantic information with dataset biases, as the content of the subsequent text is decided by the semantics of the preceding text (i.e., ``\emph{causal}''), and such relationship exists in all corpora (i.e., ``\emph{invariant}''); on the contrary, although the subsequent text would be correlative to dataset bias, such correlation changes upon different datasets.
% Specifically, since the semantic information is causally related to the subsequent text, if the model can capture the semantic information to obtain hidden states, then these hidden states would keep invariantly predictive for all instances. Hence, on the contrary, if we could find ``counter example'' on which the hidden states obtained by LLMs are predictive but not invariantly predictive, we can identify a set of counter examples characterizing LLMs' biases. 
Given the biased instances, a set of explainable bias patterns is further induced, and we devise a cost-effective and efficient in-context learning (ICL) based method to regularize LLMs using the explainable bias patterns. 

Based on the method of this paper, we construct a Python package to facilitate the automatic identification of dataset bias on Instruct Tuning Datasets. We attempt to discover biased instances and explainable biased patterns from several commonly used instruct-tuning datasets. The code is publicly available at https://github.com/spirit-moon-fly/CAL.
%. Then we select counter example pairs that are more informative for the following procedure of bias pattern induction by proposing some query strategies.
% Hence, as the combination of the causal mechanism and the active learning, CAL employs the LLM as an agent to automatically identify the dataset biases in a self-motivated manner.

%Moreover, we propose a cost-effective and efficient in-context learning (ICL) based method to prevent LLMs from utilizing dataset biases for making inferences. Concretely, in a zero-shot setting, we use the automatically induced bias patterns to explicitly tell the LLM what kind of information it should not use during inference. In the few-shot setting, we propose a counterfactual ICL method, which provides them with automatically derived counterfactual examples to balance the unintentional correlation of dataset biases with the inference target. So that LLMs can be guided towards a more unbiased mode of response.

Experimental results show that our approach can automatically induce various explainable bias patterns (some of them may be unreported), and improve the generalizability and safety of LLMs by using the ICL-based debiasing method based on the bias patterns and biased instances. 
\begin{figure*}
    \centering
    \includegraphics[width=0.99\linewidth]{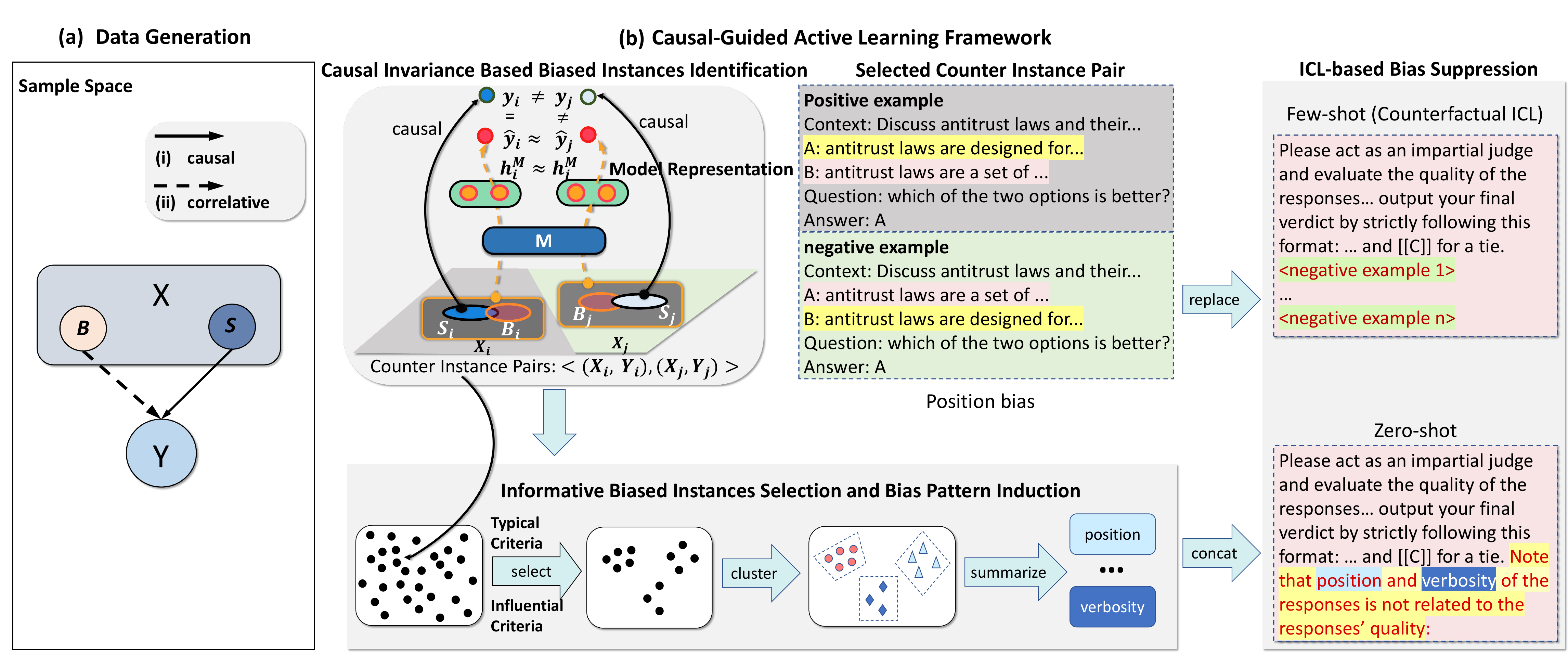}
    %\includegraphics[width=0.99\linewidth]{pic2.png}
    % % \vspace{-0.2cm}
    % % \vspace{-0.1cm}
    \caption{ (a) Dataset bias under causal perspective (b) Illustration of the Causal-Guided Active Learning framework.}
    \label{fig:archi}
\end{figure*}

% % \vspace{-0.2cm}
\section{Preliminary}
%We first analyze the stable NLU process from a causal perspective. 
% position bias做例子
% semantic relationship vs dataset biases
% 是一个映射
% X-->Y
% f_S(X)-->S
% f_B(X)-->B
% Y=P(S, B)
% % \vspace{-0.2cm}
\subsection{Dataset Bias within Textual Corpus under Causal Perspective} 
\label{sec:datsetbias}
% % \vspace{-0.1cm}
Text records and reflects the thoughts of human beings. Inherent biases such as gender and racial biases persist in the human mind, and thus are also reflected in various corpora \cite{schick2021self,navigli2023biases}. Due to potential annotation artifacts, various biases such as position and verbosity biases still broadly exist in task-specific datasets. 

Formally, as shown in Figure~\ref{fig:archi}~(a), given a piece of text $X$, the subsequent text $Y$ within a corpus $\mathcal{D}$ would be affected by two factors: (1) The semantic relationship between $X$ and $Y$, (2) The existence of dataset bias within $\mathcal{D}$. For example, given $X = \textbf{The physician hired the secretary because}$, due to the existence of gender bias, the following text $Y$ in the corpus would more likely be $\textbf{he}\ \text{was overwhelmed with clients}$, rather than $\textbf{she}$. Such $\textbf{biased relationship}$ characterizes the unwanted correlation between the context brought by dataset bias. In the following sections, for clarity, we denote the semantic relationship as $f_S(\cdot)$, and denote the biased relationship as $g_B(\cdot)$. Hence, given $X$, the conditional distribution of $Y$ given $X$ in corpus $\mathcal{D}$ can be formalized as $P(Y|X)=P(f_{S}(X), g_{B}(X)|X)$. 

The key difference between the semantic relationship and the biased relationship is that the semantic relationship possesses the \emph{causal invariance}, while the biased relationship does not. Specifically, for all instances upon all datasets, given preceding text $X$, the subsequent text $Y$ would be determined by the semantic relationship \cite{pearl2000models,pearl2009causality}, while the biased relationship only describes certain superficial statistical correlation between $X$ and $Y$. 
Consider the example where an LLM acts as a judge to assess the responses of two AI assistants, as illustrated in Figure~\ref{fig:archi}~(a): The answer ($Y$) is determined by the semantic relationship between the prompt $X$ and answer $Y$. While in the corpus, certain biases such as the position of the responses that show a correlation with the answer can be predictive. However, $Y$ is not determined by the bias and such a correlation may fail to be predictive in other instances. Hence, as $Y$ is determined by $X$, their semantic relationship is a ``causal'' relationship and invariant upon all instances. While the biased relationship is only correlative.
% % \vspace{-0.3cm}
\subsection{Biases of Generative LLMs} 
% % \vspace{-0.2cm}
During the pretraining and task-specific supervised fine-tuning process, the training objective of generative LLMs is consistent, i.e., learn to generate the subsequent text $Y$ given input text $X$. Given $X$ in corpus $\mathcal{D}$, the distribution of $Y$ can be formalized as $P(Y|X)=P(f_{S}(X), g_{B}(X)|X)$, the generative LLMs would inevitably be trained to model both $f_{S}(X)$ and $g_{B}(X)$. \
%In other words, LLM $\mathcal{M}$ would learn to capture the biased in $X$.
Therefore, given preceding text $X_i$, LLMs would not only attend to the semantics of $X_i$ but also would attend to the biased patterns such as negation word, gender indicator, position of choices, etc, to generate $Y$. As a result, during inference, the model generation $\hat{Y}$ would inevitably be affected by the dataset biases. For brevity, we denote the semantic information within $X_i$ as $S_i$ and denote the biased patterns as $B_i$. 

\subsection{Active learning} 
% % \vspace{-0.2cm}
Active learning aims at selecting the most informative instances, and then querying external information source(s) to label these data points \cite{NIPS1994_7f975a56,zhang-etal-2022-survey}. The key of active learning lies in how to devise query strategies to select the most informative instances \cite{al2022survey}. For example, uncertainty-based active learning methods aim at finding the most uncertain instances, and then send them to annotators for labeling \cite{al2021survey}. In this paper, under the automatic debiasing scenario, two key issues are: (1) finding \emph{which instance contains bias}; (2) finding the most informative biased instances. Hence, we propose a causal-guided active learning framework, which identifies the biased instances under the guidance of causal-invariance-based criterion, and finds the most informative biased instances by identifying the instances on which dataset biases have most influence on the generation of LLMs. 

% % \vspace{-0.2cm}
\section{Methodology}
% % \vspace{-0.2cm}
As Figure~\ref{fig:archi}~(b) shows, CAL contains two main components: (i) causal invariance-based biased instance identification; (ii) typical biased instances selection and bias pattern induction. Given the recognized bias patterns, we propose an in context learning-based debiasing method for regularizing LLMs. 

% % \vspace{-0.2cm}
\subsection{Causal Invariance Based Biased Instances Identification}

% 目的是要找不具备invariant predictive relationship的features。怎么找？通过找反例，构造反例。反例的存在性。
%主动学习的对应关系：标注：causal invariance criterion based 
% selection informative
We first identify a set of biased instances that reflect the inherent biases within LLMs using the difference between semantic information and biased information in the perspective of causal variance.

Compared to semantic information, the essential characteristic of biased information is that $B$ does not have an invariant causal relationship with the subsequent text, which enables the disentanglement of biased information with semantic information. Moreover, note that, the generative LLMs would capture biased information to obtain the representations (e.g. the hidden states) of input texts. Hence, \textbf{if we can find the instances where the model obtains representations that are not invariant predictive}, then the representations of these instances would contain biased information, which indicates that these instances are very likely to contain bias and could be identified as biased instances. 

%then we could identify the biased features from the representations of these instances.
% \subsubsection{Causal Invariance Criterion Based Biased Instances Identification }
Specifically, as described in Sec.~\ref{sec:datsetbias}, since the input preceding text $X$ consists of both the semantics $S$ and dataset biases $B$, hence, for an arbitrary instance $(X_i, Y_i)$ within a large enough dataset, there could exist other instance(s) $(X_j, Y_j)$, which has the following relationship with $(X_i, Y_i)$: $(B_i, S_i) \subset X_i, (B_j, S_j) \subset X_j, B_i=B_j, S_i \neq S_j$. In other words, this pair of instances shares almost the same kind of dataset biases, while the semantic information entailed in the input text is different. %Furthermore, the gold subsequent text I
The existence of such instance pairs enables the identification of biased instances using causal invariance. 
%Since the LLM $M$ would capture dataset bias to derive hidden states, then on such instance pairs, if it is probable that If we can find instance pairs 

Under such assumption, considering an instance pair \small$\langle (X_i, Y_i), (X_j, Y_j) \rangle$\normalsize, if $\mathcal{M}$ has mainly captured the semantic information $S_i$ and $S_j$, and $H^{\mathcal{M}}_i$ is close to $H^{\mathcal{M}}_j$, then $S_i$ is similar to $S_j$, so that $Sim(Y_i, Y_j)\rightarrow 1$. In other words, the LLM has captured invariant predictive information for making generations.    %according to the invariance described in Eq.~\ref{eq:ci}, 
%if an LLM $\mathcal{M}$ only captures $S_i$ to derive hidden states $H^{\mathcal{M}}_i$, then $H^{\mathcal{M}}_i$ would have the causal invariance:
% % \vspace{-0.5ex}

%\begin{shrinkeq}{-1ex}
%\small
%\begin{equation}
%\forall (X_i, Y_i) \in \mathcal{D}_j, S_i \subset X_i: Sim(Y_i, \hat{Y}_i)\rightarrow 1,\\ 
%\hat{Y}_i=u(H^{\mathcal{M}}_i),  
%\label{eq:ci}
%\end{equation}
%\end{shrinkeq}
%where $u(\cdot)$ is the function used by LLMs to generate the subsequent text based on $H^{\mathcal{M}}_i$. $Sim(\cdot)$ is a score function for evaluating if the generation of LLM $\hat{Y}_i$ is close enough to true subsequent text $Y_i$.
%Thus, for an instance pair\small$\langle (X_i, Y_i), (X_j, Y_j) \rangle$\normalsize, if $\mathcal{M}$ has captured the semantic information $S_i$ and $S_j$, and $H^{\mathcal{M}}_i$ is close to $H^{\mathcal{M}}_j$, then $Sim(Y_i, \hat{Y}_i)\rightarrow 1$, $Sim(Y_j, \hat{Y}_j)\rightarrow 1$. In other words, the LLM has captured invariant predictive information for making generations. 

\noindent\textbf{Instances on which the model fails to capture invariant predictive information}

Hence, on the contrary, if we can find an instance pair \small$\langle (X_i, Y_i), (X_j, Y_j) \rangle$\normalsize, on which $H^{\mathcal{M}}_i$ is close to $H^{\mathcal{M}}_j$, 
whereas $Sim(Y_i, Y_j)$ is low, 
%whereas $Sim(Y_i, \hat{Y}_i)$ or $Sim(Y_j, \hat{Y}_j)$ is low, 
then \small$\langle (X_i, Y_i), (X_j, Y_j) \rangle$\normalsize \thinspace can be regarded as instances on which $\mathcal{M}$ violates the causal invariance, and such instance pair can be utilized for characterizing the biases captured by LLMs. For clarity, we define such an instance pair \small$\langle (X_i, Y_i), (X_j, Y_j) \rangle$\normalsize \thinspace as a \emph{counter example pair}:

\vspace{+0.1cm}
\noindent \textbf{Definition 1} (Counter Example Pair): $\forall (X_i, Y_i)$, $(X_j, Y_j)$ $\in \mathcal{D}$, $i \neq j$, if:
\begin{shrinkeq}{-1ex}
\begin{small}
\begin{equation}
S(H^{\mathcal{M}}_i, H^{\mathcal{M}}_j)\!>\!\tau, \\ 
%s.t. \ Sim(Y_i,\hat{Y}_i)\!<\!\alpha \ \vee \ Sim(Y_j,\hat{Y}_j)\! <\! \alpha,
s.t. \ Sim(Y_i,Y_j)\!<\!\alpha,
\label{eq:def}
\end{equation}
\end{small}
\end{shrinkeq}
\unskip
where $\mathcal{D}$ is the dataset, $S(\cdot)$ is a score function measuring the similarity between \small $H^{\mathcal{M}}_i$\normalsize and \small $H^{\mathcal{M}}_j$\normalsize, $\tau$ is a threshold controlling the confidence that \small $H^{\mathcal{M}}_i$ \normalsize and \small $H^{\mathcal{M}}_j$ \normalsize can be regarded as close enough, and $\alpha$ is another threshold ensuring that 
%$Y_i$ and $\hat{Y}_i$, $Y_j$ and $\hat{Y}_j$ 
$Y_i$ and $Y_j$
can be regarded as sufficiently different.

Definition 1 enables us to detect all counter example pairs within the dataset $\mathcal{D}$. On these counter example pairs, the invariance is violated so that subsequent texts are generated based on biased information. Hence, $H^{\mathcal{M}}_i$ and $H^{\mathcal{M}}_j$ contains the bias information $B_i = B_j$. However, the aforementioned theory is built upon the assumption that LLMs have captured the predictive information (including bias and semantic information). In fact, when $X_i$ is very difficult or ambiguous, it cannot be ruled out that the LLM does not capture any predictive information. To rule out such instances, we introduce an additional filtering process using a \textbf{Predictive Criterion}, which requires that $\mathcal{M}$ should at least make a proper generation for the instance $i$ or $j$, since if on both $i$ and $j$ model generation are improper, it is rather probable that $\mathcal{M}$ has not captured any predictive information in $X_i$ or $X_j$:
\vspace{-0.4cm}

\begin{shrinkeq}{-1ex}
\begin{small}
\begin{align*}
%\textbf{Predictive Criterion}: \\ 
\tag{3} Sim(\hat{Y}_i,Y_i)\! >\! \beta \vee Sim(\hat{Y}_j,Y_j)\! >\! \beta,
\end{align*} 
\end{small}\end{shrinkeq}
where $\hat{Y}_i$, and $\hat{Y}_j$ are the generated subsequent text, 
%$l_j$ is the index of correct label, $\hat{p}_{i,l_j}$ is the predicted probability of the gold subsequent text, 
$\beta$ is a threshold ensuring that $\hat{Y}_i$ and $Y_i$ can be regarded as similar enough so that $\hat{Y}_i$ can also be seen as a correct answer (the same for $\hat{Y}_j$).
%and $\tau_p \in [0,1]$ is a threshold.
%This is because, if $\mathcal{M}$ fails to correctly generate the subsequent text for both $X_i$ and $X_j$, it is rather probable that $\mathcal{M}$ has not captured any information in $X_i$ or $X_j$ related to $Y_i$ and $Y_j$, so that it is unsuitable for characterizing the biases within LLMs. 

%Typical and informative --> 对应多样性
% % \vspace{-0.3cm}
\subsection{Selection of Informative Biased Instances and Bias Pattern Induction}
\label{sec:selection}
% % \vspace{-0.2cm}
%Typical bias instances which is informative for the procedure of bias pattern induction should have the following features
% we considering the characteristics that a informative counter example pair should possess.
Using the criterion mentioned above, we could identify a set of instances that contain bias (i.e., counter instance pairs) as they violate the causal invariance criterion. Next, we hope to select a subset that is more informative and contains typical dataset bias. So that we can further induce explainable patterns of biases to prevent the LLMs from utilizing bias. To this end, we consider that:

\noindent\textbf{Typical Biased Instances Identification}
Firstly, for any input text $X_i$,
%consider a scenario where an LLM generates subsequent text that significantly diverges from the accurate or intended subsequent text. 
if the probability that $Y_i$ is properly generated is rather low, it suggests that biased information significantly hinders the LLM. Hence, such examples would contain a high level of bias and could be informative biased instances.

Secondly, for a counter instance pair \small$\langle (X_i, Y_i), (X_j, Y_j)  \rangle$\normalsize, if the corresponding generation of LLM $\hat{Y}_i$ and $\hat{Y}_j$ is rather different, it means the influences of dataset bias are diversified and hence it would be challenging to summarize a unified bias pattern based on these samples. Conversely, if $\hat{Y}_i$ and $\hat{Y}_j$ are similar, it would be easier to conclude the influence caused by the bias, as the influence of dataset bias is typical. 
Based on the two characteristics, we introduce the following two criteria to select the informative biased instances:
\vspace{-1ex}

\begin{shrinkeq}{-1ex}
\begin{small}
\begin{flalign}
&\textbf{Influential Criterion}\!: 
\hat{p}_{j,l_j}\!<\!\tau_p,\!\ s.t. \ Sim(\hat{Y}_j,\!Y_j)\!<\!\alpha, \\ 
&\textbf{Typical Criterion}\!:  
Sim(\hat{Y}_i,\!\hat{Y}_j)\!>\!\beta,
\end{flalign}\end{small}\end{shrinkeq}
where $l_j$ is the gold subsequent text, $\hat{p}_{i,l_j}$ is the predicted probability of gold subsequent text, and $\tau_p \in [0,1]$ is a threshold for controlling the probability that $\mathcal{M}$ generates gold subsequent text.

\noindent\textbf{Bias Pattern Induction}
Based on the identified informative biased instances, we further induce certain explainable patterns that characterize several major types of dataset biases among the corpus.
To this end, we first group the counter example pairs into several clusters, and then induce patterns for each cluster. 

%negative example和positive example得说明一下
The cluster of counter example pairs is derived based on the \emph{bias representation vectors} of the counter example pairs, which refers to the representation vector of the bias component of a counter example pair. We obtain the bias representation vectors of a counter example pair\small$\langle (X_i, Y_i), (X_j, Y_j) \rangle$\normalsize by extracting the \emph{similar parts in the representations of two examples} (i.e. $H^{\mathcal{M}}_i$ and $H^{\mathcal{M}}_j$). This is because, as described in the definition of counter instance pair, the similar parts of $H^{\mathcal{M}}_i$ and $H^{\mathcal{M}}_j$ carry the biased information. 
%Additionally, the \emph{typical criterion} ensures that similar parts between $H^{\mathcal{M}}_i$ and $H^{\mathcal{M}}_j$ primarily encompass bias. Based on the $\texttt{typical criterion}$, we know that all the selected counter example pairs contain one example on which LLMs generate incorrect subsequent text (we refer to this example as the negative example and the other example in the counter example pair as the positive example for convenience). If the negative example primarily contains semantic information, its probability of generating the correct subsequent text would be high. This would contradict criterion 2 which requires the probability of generating the gold subsequent text would be low, implying that the negative example mainly consists of bias and contains relatively limited semantic information. Consequently, the similar parts of the representations corresponding to these two examples mainly contains bias. 
% Hence, if we can derive the vector representation of each instance within a counter instance pair and extract their common components, then we can derive the \emph{bias representation vectors} of the counter example pairs.

After obtaining the representation vector of the biases in each counter example pair, we first apply Principal Component Analysis to reduce the dimension of bias representation vectors to two dimensions. As the dimension of data increases, the distances between data points become increasingly similar, so traditional distance metrics (such as Euclidean distance) would be less effective and in turn affect the performance of clustering algorithms. Then we perform clustering based on the dimension-reduced biased representation vectors using the density-based clustering method DBSCAN \footnote{We find that the first two principal components can explain over 96\% of the total variance. Thus, the left part may mainly be noise and would disturb the process of clustering. Hence, we perform the DBSCAN based on the first two PCA components.
}. Finally, we obtain counter example pairs within each cluster, and provide them to GPT 4 for summarizing bias patterns. For example, from the selected counter example pair in Figure~\ref{fig:archi}~(b), we can summarize the position bias. 
% % % \vspace{-0.3cm}
\subsection{In Context Learning-based Bias Suppression}
% % % \vspace{-0.2cm}
To prevent the LLMs from utilizing dataset biases for making generation, meanwhile avoiding the drawbacks of fine-tuning-based methods, we propose a cost-effective and efficient in-context learning (ICL) based method. Concretely:

In the \textbf{zero-shot} scenario, as shown in Figure~\ref{fig:archi}~(b), we use the automatically induced bias patterns to explicitly tell the LLM what kind of information it should not use during inference 
by appending the text \colorbox{lightgray}{``\emph{[bias xxx]} is not related to \emph{[the goal of the task]}''} to the end of the original prompt. 
%For the example in Figure~\ref{fig:archi}~(b), in the scenarios of LLM as a judge to evaluate different AI assistants' response, we have summarized the bias position bias and verbosity bias. Then by appending the text ``\emph{position bias and verbosity bias} is not related to \emph{the responses’ quality}'' to the end of the original prompt, we inform the LLM not to judge the relationship between the premise and the hypothesis based on position bias and verbosity bias to debias it.

In the \textbf{few-shot scenario}, we propose a counterfactual ICL method, which provides LLMs with automatically derived counterfactual examples to correct the LLM's belief about bias. Specifically, if we could find ``counterfactual examples'', on which using biased information for inference would conversely lead to incorrect generations. Then by providing such examples to LLMs in the prompt, LLMs would be implicitly informed that the biased information is not related to the subsequent text, and thus it would be regularized to not use biased information for making inferences. To find such ``counterfactual examples'', notice that according to the Influential Criterion, for an arbitrary counter example pair\small$\langle (X_i, Y_i), (X_j, Y_j) \rangle$\normalsize, the LLM would make improper generation upon instance $i$ or $j$. Without generality, we denote this instance as $i$ and instance $i$ could be regarded as a counterfactual example for debiasing LLMs. Intuitively, in instance $i$ the dataset bias leads to improper generations, which is contrary to most cases within the corpus, hence we call instance $i$ as a counterfactual example.  
% Such there exists an example that LLM
% the positive example and negative example shares the same bias information $B_i = B_j$, while their semantic information are different $S_i \neq S_j$.
% So we can consider the negative example as a ``counterfactual example'' of the positive example which is obtained by using \textbf{do-operation} to change the semantic information of the positive example. Thus, we can automatically derive counterfactual examples by collecting negative examples in the counter example pairs.

%To balance the unintentional correlation of dataset biases with the inference target,
Hence, to correct the LLM's belief about bias, we construct the prompt with such counterfactual examples using the following format: \colorbox{lightgray}{``<EXAMPLES>. Note that you should not utilize} \colorbox{lightgray}{biased information to make generations''}, where <EXAMPLES> are the counterfactual examples. 
%According to the [input text] of the counterfactual example, the model would generate subsequent text $t$ based on the LLM's inherent belief about bias. However, the LLM also observes that the text generated without utilizing bias is entirely different from $t$. As a result, the model learns to some extent that it should not employ its inherent belief about bias during generation, thereby correcting the LLM's belief about biases.
%% % \vspace{-0.2cm}
\section{Experiments}
%% % \vspace{-0.2cm}
\subsection{Experimental Details}
%% % \vspace{-0.1cm}
In this work, we use llama2-13B-chat \cite{touvron2023llama} and vicuna-13B-v1.5 \cite{chiang2023vicuna} for our experiments. Without loss of generality, we examine our approach on datasets that have a clear set of possible answers, e.g., multiple-choice question-answering task. So that we can implement the $Sim(\cdot)$ function in Equation~\ref{eq:def} using an exact match of strings. If matched, the function's value is 1, otherwise it's 0. So $\alpha$ and $\beta$ can be any value between 0 and 1. 
%Additionally, the cosine function between the representations of two input texts is employed as the scoring function $S(\cdot)$ to measure the similarity between these input texts. For convenience, we use the embedding vector of the last token at the top of the LLM's layer to derive the representations of input texts using llama2-13B-chat and vicuna-13B-v1.5 respectively.
Additionally, we derive the representation of input text by employing the embedding vector of the last token at the top of the LLM's layer, and the cosine function is employed as the scoring function $S(\cdot)$ to measure the similarity between these hidden states.

To derive bias representation vector of a counter example pair, we need to extract similar parts in the hidden states corresponding to two examples of the counter example pair. This is because, the similar parts in the hidden states carry the biased information as mentioned in Section~\ref{sec:selection}. To this end, we obtain the similar components of two hidden states in an element-wise manner. Specifically, we use the following function:
\vspace{-0.2cm}

\begin{shrinkeq}{-1ex}
\begin{small}
\begin{equation}
f(H_{ik},H_{jk})\!=\!\left\{
\begin{matrix}
(H_{ik}\!+\!H_{jk})/2 & if \frac{|H_{ik}-H_{jk}|}{H_{ik}+H_{jk}}\!<\!\mu
\\ 0 & otherwise
\end{matrix}
\right.
\end{equation}
\end{small}
\end{shrinkeq}
\unskip
where $H_{ik}, H_{jk}$ are the k-th element of two hidden states $H^{\mathcal{M}}_i$ and $H^{\mathcal{M}}_j$. 
%and $\mu$ is a threshold for controlling that the two elements of certain position can be considered as similar enough. 
If $H_{ik}$ and $H_{jk}$ are similar enough, then their difference should be relatively small. We measure such difference using $|H_{ik}-H_{jk}|$/$|H_{ik}+H_{jk}|$, and then using a threshold $\mu$ to determine if $|H_{ik}-H_{jk}|$/$|H_{ik}+H_{jk}|$ is small enough, in other words, $H_{ik}$ and $H_{jk}$ are similar enough. If they are similar enough, we use the average of $H_{ik}$ and $H_{jk}$ to represent the k-th element of the bias representation vector of a counter example pair. If not, we use 0 to represent the k-th element of the bias representation vector.
In practice, we choose $\mu$ by controlling the ratio that the two elements of certain position can be considered as similar enough in MNLI dataset when using llama2-13B-chat. We set a strict threshold of 0.15 for the ratio to ensure that the bias representation vectors of the counter example pairs have purer bias information. 

Moreover, note that, it is UNNECESSARY to run CAL upon the whole corpus to obtain the biased instances and the bias patterns. A subset would be enough (e.g., 2,0000 instances) to save the computational cost. In the Section~\ref{sec:size} and Appendix ~\ref{sec:sensitivity}, we provide the sensitivity analysis of the dataset size and hyperparameters, including the influence of the size of data for obtaining the counter instance pairs and the thresholds.
%we first calculate element-wise similarity between two corresponding elements in two hidden states by computing the ratio of the difference to the sum of these two elements and design a similarity threshold. If the similarity is higher than the threshold, we use the  
%During bias pattern induction, we summarize three bias patterns using GPT-4 from each cluster category.
%In zero-shot scenarios, we discovered that providing debiasing prompt containing more than two bias patterns may lead to a decline in performance, even if using any of these bias patterns individually results in a performance increase. Hence, in the debiasing prompt, we use the first two bias patterns obtained from the cluster category with the highest number of counter example pairs because they can represent the most common bias. 

In few-shot scenarios, to make results comparable, we ensure that the number of examples in prompts equals that used in other few-shot baselines. 
Additionally, we maintain the order of gold answers that appear in the few-shot examples to avoid introducing additional label bias. 
We report the average results across 10 runs considering the randomness in sampling counterfactual examples.

Below we call our method zero-shot-CAL and few-shot-CAL in zero-shot and few-shot settings respectively. More details about experimental settings are provided in Appendix.
%In threspectivele process of selectinttings g each few-shot example, we need to choose more representative counterfactual examples due to the limited number of few-shot samples we can select. Specifically, we select the counterfactual example with the highest predicted probability for its generated answer. This is because the LLM's confidence in generating its answer indicates a greater influence of the bias contained in this counterfactual example, making this counterfactual example more representative. Additionally, we also report the average results across 10 runs by removing the condition for selecting representative examples.  

%% % \vspace{-0.3cm}
\subsection{Evaluation Tasks}
%% % \vspace{-0.2cm}

We examine the effectiveness of CAL by investigating whether CAL could debias LLMs to improve the generalizability and unharmfulness of LLMs. 

To evaluate the improvement of generalizability, we conduct experiments by deriving biased instances and bias patterns on dataset A and utilizing the identified instances and biased patterns to debias both dataset A and dataset B. Heuristically, two datasets A and B may share different dataset bias distributions. If an LLM only adapts to dataset A, then its performance upon dataset B would be impacted. On the contrary, if an LLM can focus more on semantics, the performance on both datasets would be improved. 
Hence, the generalizability could be evaluated by \emph{the performance improvement compared to baseline methods}. %相比于基线分布外性能一致提升，则说明
Specifically,  We evaluate our approach on benchmarks representing two categories of bias: (1) Generative-LLM-specific biases. We employ the Chatbot and the MT-Bench datasets \cite{zheng2023judging} as benchmarks.
On both datasets, LLM is required to choose a better response from two candidates.
%, and \citet{zheng2023judging} shows that LLM will exhibit the same biases such as position bias and verbosity bias when making choices on both datasets. 
We induce the bias patterns on the Chatbot dataset, then test whether the Chatbot-based bias patterns can be utilized to debias LLMs on both the Chatbot and the MT-Bench dataset. (2) Task-specific biases. We choose the natural language inference dataset MNLI \cite{williams2018broad} and the corresponding manually debiased dataset HANS \cite{mccoy2019right} as benchmarks. Hence, models that only utilize the biased information often perform close to a random baseline on HANS. 
The bias patterns are induced from the MNLI dataset, then test whether CAL can utilize the induced bias patterns to debias LLMs on both the MNLI and the HANS datasets. 

To evaluate the improvement of unharmfulness, we conduct experiments on the BBQ \cite{parrish2022bbq} and the UNQOVER \cite{li2020unqovering} dataset, which is designed for evaluating stereotype biases (such as gender bias and racial bias) of LLMs. These two datasets containing 9 and 4 types of stereotype bias, respectively. On these two datasets, if the model achieves a higher accuracy, then it could be regarded as having a lower likelihood of containing stereotypes.

On Chatbot and MT-Bench dataset, model performance is evaluated based on the agreement ratio between human-majority annotations and LLMs. On other datasets, model performance is evaluated using accuracy. 

%We examine the generalizability of our method on OOD samples by comparing the performance on different datasets that shares the same task but is not used for extracting biased instances. To evaluate our methods' ability to guide LLMs toward a more unharmful generation mode, we conduct experiments on datasets that is desinged for evaluating stereotype biases of LLMs. 

\begin{table}[tbp] 
\centering
\small
\setlength{\tabcolsep}{1mm}{
\begin{tabular}{ c|c c|c c|c c} 
\toprule %添加表格头部粗线
\multicolumn{1}{c}{}&\multicolumn{4}{c}{\textbf{Generalizability Evaluation}} &\multicolumn{2}{c}{\textbf{Unharmful E.}}   \\ 
\cmidrule(lr){2-7}
\multicolumn{1}{c}{LLAMA2}&Chatbot&MT &MNLI&HANS &BBQ&UQ   \\ 
\hline %绘制一条水平横线
\specialrule{0em}{1.5pt}{1.5pt}
\multicolumn{1}{c}{ZS}      &38.9         &34.5 &65.9&52.9 &47.6&23.4              \\
\multicolumn{1}{c}{ZS-known}&\textbf{42.7}&41.2 &67.2&55.0 &51.1&59.4              \\
\multicolumn{1}{c}{FS}      &40.4         &46.9 &66.1&53.1 &49.5&23.1              \\
\multicolumn{1}{c}{ZS-CAL}  &40.5         &43.3 &\textbf{67.4}&55.5 &51.5&\textbf{60.3}  \\
\multicolumn{1}{c}{FS-CAL}  &41.6&\textbf{49.8} &64.1&\textbf{59.3} &\textbf{53.5}&32.3     \\
\midrule
\multicolumn{1}{c}{Vicuna}&Chatbot&MT &MNLI&HANS &BBQ&UQ   \\ 
\hline %绘制一条水平横线
\specialrule{0em}{1.5pt}{1.5pt}
\multicolumn{1}{c}{ZS}        &35.2&43.8         &66.7&38.3 &47.9&33.3     \\
\multicolumn{1}{c}{ZS-known}  &38.2&\textbf{50.0}&69.6&55.0 &49.5&35.2     \\
\multicolumn{1}{c}{FS}        &37.3&46.9&\textbf{71.0}&62.5 &59.7&48.9     \\
\multicolumn{1}{c}{ZS-CAL}    &\textbf{39.9}&\textbf{50.0}&69.8&57.1 &48.5&35.3     \\
\multicolumn{1}{c}{FS-CAL}   &39.8&49.4&69.5&\textbf{63.7} &\textbf{65.5}&\textbf{58.5}   \\
\bottomrule %添加表格底部粗线
\end{tabular}
}
\caption{Comparison of CAL with baselines in both zero-shot and few-shot settings across two LLMs. ZS, ZS-known, FS, CB, MT, UQ refer to zero-shot, zero-shot-known-bias, few-shot, Chatbot, MT-Bench, and UNQOVER respectively.}
\label{tab:main}
\end{table}

%% % \vspace{-0.3cm}
\subsection{Baseline Methods}
%% % \vspace{-0.2cm}
We compare the casual-guided active learning method with two categories of baseline methods:

\noindent\textbf{vanilla zero-shot and few-shot baselines} 
We examine the vanilla zero-shot and few-shot performance of LLMs using the prompt of \citet{zheng2023judging,si2023prompting,xu2023anllm}. 

\noindent\textbf{zero-shot-known-bias}
These methods mainly rely on human prior knowledge of bias to design debiasing prompts.
For Chatbot and MT-Bench datasets, we compare CAL with the debiasing method of swapping positions proposed in \citet{zheng2023judging}. For BBQ and UNQOVER datasets, we follow the instruction from \citet{si2023prompting} to avoid stereotype bias. For MNLI and HANS datasets, we use the debiasing prompt to prevent lexical overlap and subsequence bias proposed in \citet{mccoy2019right}.

To the best of our knowledge, the only few-shot debiasing method comes from 
\citet{oba2023contextual}. However, this method is unsuitable for our dataset.  Details can be seen in Appendix~\ref{sec:baselinedetail}.

% \vspace{-0.2cm}
\subsection{Main Results}
% \vspace{-0.1cm}
We list the experimental results of two LLMs on six datasets in Table~\ref{tab:main}. From which we find that:

\textbf{(1)} Compared to the vanilla zero-shot shows that, in general, the prior knowledge-based zero-shot debiasing methods show improved performance on all the datasets. This indicates that through ICL, LLMs can both effectively debias themselves and avoid the in-distribution performance degradation which is always associated with fine-tuning-based approaches \cite{du2023towards}, suggesting the superiority of ICL-based debiasing methods.

%Comparing the automatic debiasing methods with the prior knowledge based debiasing methods shows that, in general, the prior knowledge based methods still show better performance on both in-distribution test sets and OOD challenge sets. This is because the distribution of biases in NLU datasets can be rather complex, bringing challenges for precisely and comprehensively detecting the potential biases.
\begin{figure}
    \centering
    \includegraphics[width=1.0\linewidth]{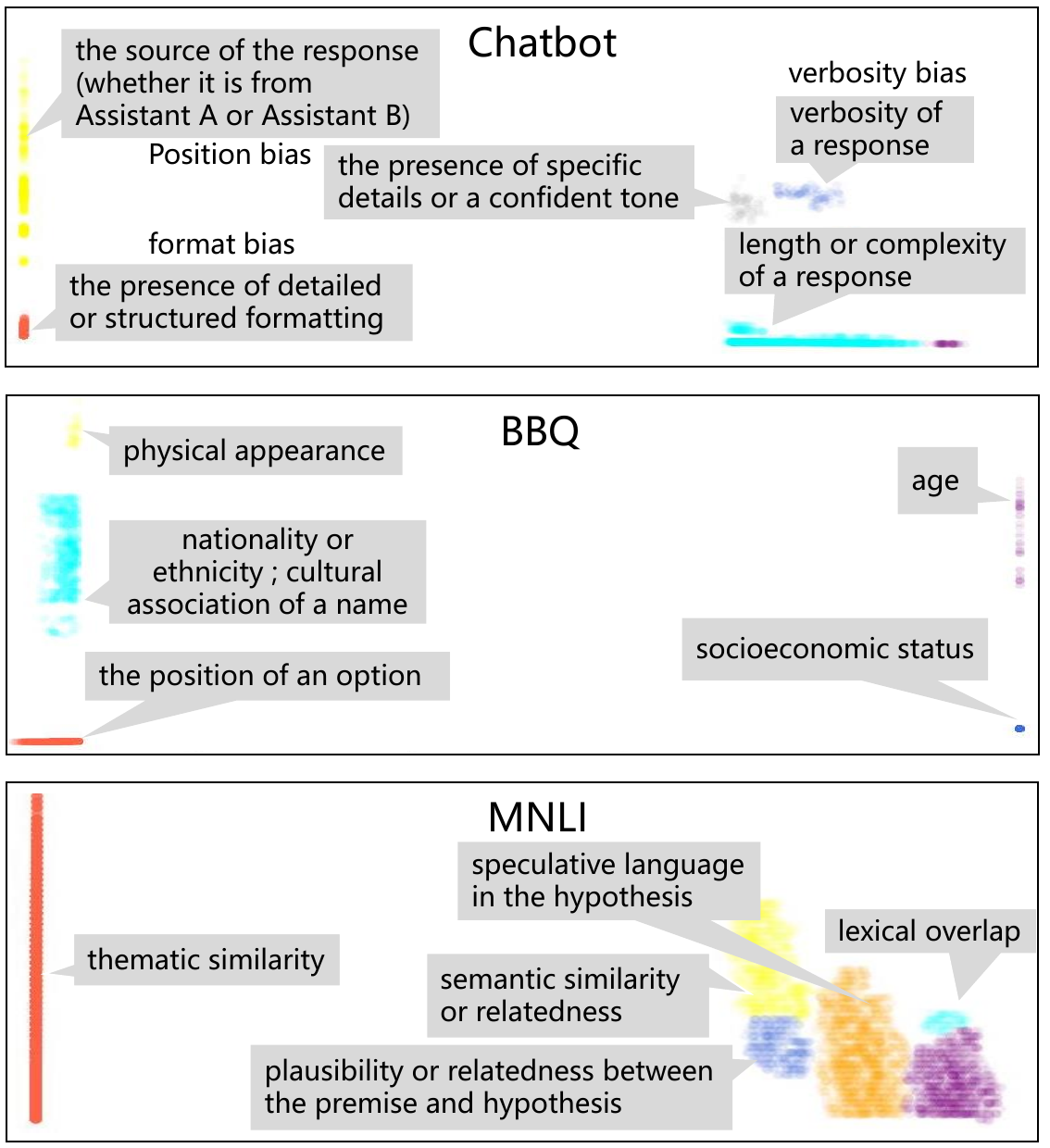}
    \caption{Results of bias pattern induction. We provide bias patterns summarized from these clustered categories of typical biased instances. }
    \label{fig:bias_patterns}
\end{figure}

\textbf{(2)} Compared to the zero-shot baselines and few-shot baselines, in general, few-shot CAL achieves consistent performance improvement on the two categories of benchmarks. This demonstrates that, CAL can improve both the generalizability and the unharmfuless of LLMs, and suggests that by utilizing the essential differences between semantic information,
%and dataset bias and devising some selection strategies
CAL can identify a set of biased instances, and the counterfactual ICL-based prompts can effectively leverage the biased counterfactual examples to debias LLMs. 
%Besides, these results also suggest that it is feasible to automatically derive counterfactual examples from existing datasets. This is quite exciting because synthesizing counterfactual samples through manually designing vocabularies requires extensive human labor and is dependent on prior knowledge of the information we wish to intervene in, which limits the application of counterfactual in many scenarios, such as debiasing. 

\textbf{(3)} Compared with vanilla zero-shot baselines, zero-shot CAL can consistently improve model performance on all the datasets, and even surpass the performance of few-shot methods on part of benchmarks. 
The effectiveness of zero-shot CAL suggests that the biased patterns induced by CAL are typical and truly exist in the datasets. This is because, by utilizing the causal invariance together with the influential and typical criterion, a set of \textbf{typical} biased instances could be selected, so that the biased patterns could be effectively induced. 
%Then by telling LLMs not to predict based on the induced biased patterns, we can surpass bias within LLMs so that improving the generalizability of LLMs.
%This indicates that our automatically derived debiasing prompts that is based on the automatically inducted bias patterns can effectively debias LLMs, which in-turn demonstrated that by utlizing the essential differences between semantic information and dataset bias and devising some selection strategies, we can select biased instances that contain different kinds of typical bias and utilizing the inductive ability of GPT-4 to summarize bias patterns from them.
% More analysis of the bias patterns we derived can be seen in \ref{generalize}. 
% 后续做实验设计和unqover格式一致的prompt

\textbf{(4)} Compared with the prior knowledge-based zero-shot debiasing methods, zero-shot CAL shows comparable or better performance on two categories of benchmarks. On the one hand, the complexity of the distribution of dataset biases brings challenges for precisely and comprehensively detecting the potential biases. On the other hand, the comparable performance between zero-shot CAL and prior knowledge-based zero-shot debiasing methods shows the effectiveness of our approach, and the potential for application in real-world scenarios, as it would be impractical to investigate all biases for various real-world corpus.
% Additionally, the automatically generated debiasing prompts tend to be worse than those designed by humans, even for the same bias pattern.
%In general, improvements on the challenge sets come with the expense of the in-distribution performance. This is because, on the one hand, the dataset biases also provide additional clues leaking the label information \cite{zhang2019theoretically,tsipras2018robustness,sanh2020learning}. Omit of the bias would naturally lead to a performance decrease on in-distribution samples. On the other hand, to improve the OOD performance, the models should identify more abundant biased features. This would increase the risk that the semantic information is mistaken as the bias information. This highlights the necessity of thoroughly disentangling the bias information with the semantic information.

%\textbf{(4)} We also devise experiments that uses the automatically derived counterfactual examples and bias patterns simultaneously and we refer to it as FS-CAL*. From Table~\ref{tab:main}, we can find that the performance of FS-CAL* is better than FS-CAL, which demonstrates that our ICL-based debiasing method's performance can be further improved.

% \textbf{(4)} Compared to baselines, our methods show better performance on both BBQ and UNQOVER datasets, which indicates that our approach can suppress stereotype bias so that LLMs are guided towards a more unharmful generation mode.

\textbf{(5)} In general, our method is effective for both llama2-13B-chat and vicuna-13B-v1.5. This suggests the prevalence of biases in LLMs, and demonstrates the generality of our approach in adapting to different LLMs.

% \vspace{-0.3cm}
\subsection{Case Study}
\label{generalize}
% \vspace{-0.2cm}
We argue that one of our potential major contributions is that by utilizing the causal invariance together with the influential and typical criterion, we can identify a set of \textbf{typical} biased instances, and then autonomously summarize explainable bias patterns from data. In Figure~\ref{fig:bias_patterns}, 
we present the results of clustering analysis based on the bias representations derived from bias instances, and bias patterns summarized from the clustered categories. Experiments are conducted using llama2-13B-chat. 

Overall, it can be observed that bias representations are concentrated in several distinct groups after dimensionality reduction through PCA. Moreover, the bias patterns summarized based on different clustering categories are also distinguished. This indicates that our method could discover different types of biased instances and then induce bias patterns.

Based on the counter example pairs derived from the Chatbot dataset, CAL can simultaneously induce position bias, verbosity bias, and format bias, which is separately identified by several previous research \cite{zheng2023judging,zhu2023judgelm}, suggesting the efficiency and effectiveness of our approach.
Furthermore, we also observe several potential bias patterns such as ``length or complexity of a response'' and ``the presence of specific details or a confident tone'', 
\begin{table}[tbp] 
\centering
\small
\renewcommand\arraystretch{1}
\setlength{\tabcolsep}{1mm}{
\begin{tabular}{c c c c c c c c c c} 
\toprule %添加表格头部粗线
\multicolumn{1}{c}{GPT-4}&Chatbot&MT &MNLI&HANS &BBQ&UQ          \\ 
\hline %绘制一条水平横线
\specialrule{0em}{1.5pt}{1.5pt}
\multicolumn{1}{c}{ZS}        &57.4&65.3&80.1&65.1&90.7&88.9     \\
\multicolumn{1}{c}{ZS-CAL}    &58.9&66.2&82.4&67.8&87.0&98.7     \\
\bottomrule %添加表格底部粗线
\end{tabular}
}
\caption{bias pattern generalization experiments}
\label{tab:generaliization}
\end{table}
that are previously unreported. 
When we tell llama2-13B-chat not to make predictions 
based on these biases, its performance increases on both Chatbot and MT-Bench datasets, suggesting that these patterns could be the truly existing biases.
Among the 9 known types of stereotype biases in the BBQ dataset \cite{parrish2022bbq}, our method can automatically identify 7 of them without prior knowledge (the bias of gender, sexual orientation, and religion are grouped into ``cultural association of a name'' during the bias induction procedure). 
On the MNLI dataset, we observe some unreported new bias patterns such as ``speculative language in the hypothesis'' (e.g., should, perhaps, possibly), 
and we can also improve the performance of llama2-13B-chat by telling it not to make predictions based on these bias patterns. 
More analyses of the counter example pairs can be seen in Appendix~\ref{sec:case}.

The automatically summarized bias patterns demonstrate the diversity of dataset biases in practical datasets, and it would be impractical to identify all of them manually. Therefore, there is an urgent need for methods to automatically identify biases. As a pioneer work, we explored that the LLMs can be automatically debiased by combining the causal mechanism and active learning, suggesting the potential feasibility of utilizing LLMs to autonomously debias themselves.
%We leave more rigorous validation of these bias patterns for the future.

% % \vspace{-0.3cm}
\subsection{Generalizablity of the Induced Bias Patterns}
% % \vspace{-0.2cm}

The pretraining corpus of different LLMs share unnegligible overlaps, so they would also possess common biases.
Hence, we investigate the generalizability of the automatically induced bias patterns by testing if it is possible to debias LLM-A based on the bias pattern identified from another LLM-B. Specifically, we attempt to debias GPT-4 based on the bias pattern (and the corresponding debiasing prompt) identified from llama2-13b-chat. Experimental results are shown in Table~\ref{tab:generaliization}, from which we can observe that compared to vanilla zero-shot, ZS-CAL achieves higher performance in most cases. This demonstrated that different LLMs might share similar bias patterns and we can debias an LLM based on the bias pattern identified from other LLMs, which further demonstrates the universality of our method.

\begin{table}[tbp] 
\centering
\small
\renewcommand\arraystretch{1}
\setlength{\tabcolsep}{1mm}{
\begin{tabular}{c c c c c c c c c c} 
\toprule %添加表格头部粗线
\multicolumn{1}{c}{}&Chatbot&MT &MNLI&HANS &BBQ&UQ          \\ 
\hline %绘制一条水平横线
\specialrule{0em}{1.5pt}{1.5pt}
\multicolumn{1}{c}{ZS}        &38.9&34.5&65.9&52.9&47.6&23.4     \\
\multicolumn{1}{c}{ZS-CAL}    &40.4&42.8&67.4&55.3&51.7&51.0     \\
\bottomrule %添加表格底部粗线
\end{tabular}
}
\caption{Using Qwen-72b-chat for bias pattern induction}
\label{tab:qwen}
\end{table}

\subsection{Influence of the Choice of Bias Pattern Induction Model}

In the above sections, we induce the explainable bias patterns using GPT-4. We also attempt to use the open-source LLM Qwen1.5-72B-Chat for inducing bias patterns to examine the generalizability. As Table~\ref{tab:qwen} shows, the results still outperform the baseline methods with the biased patterns induced by free open-source LLM, while slightly inferior to that of GPT-4. This shows the generality of our approach, and implicates the potential application in real-world scenarios.

\begin{table}[tbp] 
\centering
\small
\renewcommand\arraystretch{1}
\setlength{\tabcolsep}{1mm}{
\begin{tabular}{c c c c c c} 
\toprule %添加表格头部粗线
\multicolumn{1}{c}{}&MNLI&HANS      \\ 
\hline %绘制一条水平横线
\specialrule{0em}{1.5pt}{1.5pt}
\multicolumn{1}{c}{ZS}           &65.9&52.9&    \\
\multicolumn{1}{c}{ZS-CAL}       &67.4&55.5&     \\
\multicolumn{1}{c}{ZS-CAL(20\%)} &67.1&55.4&     \\
\multicolumn{1}{c}{FS}           &66.1&53.1&    \\
\multicolumn{1}{c}{FS-CAL}       &64.1&59.3&     \\
\multicolumn{1}{c}{FS-CAL(20\%)} &64.0&59.7&     \\

\bottomrule %添加表格底部粗线
\end{tabular}
}
\caption{Using 20\% subset for bias pattern induction}
\label{tab:subset}
\end{table}

\subsection{Influence of the Dataset size}
\label{sec:size}
To investigate the influence of the dataset size used in our framework, We conducted experiments using a 20\% subset of the MNLI dataset utilized in our main experiments, employing the llama2-13b-chat model. As Table~\ref{tab:subset} shows, the performance of CAL keeps relatively stable with 20\% data. Moreover, our approach still far outperforms the baseline method on the HANS dataset, which demonstrates the effectiveness of our approach to debias LLMs. This indicates that our method is still effective in situations where data is relatively scarce.

% % \vspace{-0.3cm}
\section{Related Work}
% % \vspace{-0.2cm}
% % \vspace{-0.01cm}
% Debiasing methods in NLP, active learning, ICL, 
% % \vspace{-0.1cm}
% \subsection{Debiasing methods in NLP}
% % \vspace{-0.2cm}
% % \vspace{-0.1cm}
Previous analyses demonstrate that LLMs still suffer from biases such as position bias \cite{zheng2023judging} and stereotyping bias \cite{shaikh-etal-2023-second}. To mitigate the LLMs' biases, one line of methods relies on researchers' prior knowledge to artificially recognize the potential dataset biases, followed by debiasing through prompt-based regularization or aligning with human through instruct tuning \cite{oba2023contextual,liu2023trustworthy,ganguli2023capacity}. However, these methods are limited by the dependence on researchers' prior. Moreover, due to the diversity of dataset biases \cite{poliak2018hypothesis,schuster2019towards,schick2021self}, it is unrealistic to identify them one by one manually. To tackle these issues, automatic debiasing methods are proposed. They automatically extract bias features characterizing the dataset biases by training certain biased models \cite{utama2020towards,du2023towards,sanh2020learning,lyu2023feature} for regularizing the main model. However, such methods are designed for discriminative models and are hard to adapt to generative LLMs. 
%Moreover, as these methods are This is because the fine-tuning-based debiasing process is computationally costly for LLMs and may come at the cost of undermining the versatility of LLMs on real-world applications. 

In this paper, we propose a causal-guided active learning framework for automatically debiasing generative LLMs. 
%Active learning aims at selecting the most informative instances and then querying external information source(s) to label these data points \cite{NIPS1994_7f975a56,zhang-etal-2022-survey}. The key of active learning is how to devise query strategies to select the most informative instances for labeling \cite{al2022survey}. In this paper, 
We borrow the idea from active learning \cite{zhang-etal-2022-survey} by first automatically identifying the potentially biased instances using the causal invariance mechanism, then automatically selecting the informative biased instances using the typical criterion and influential criterion. 
%There are many data query strategies aims at selecting the most challenging or error-prone samples to annotate \cite{al2021survey}.By utilizing the essential differences between semantic information and dataset bias and devising some selection strategies, we can actively select typical and informative biased instances. 
Based on such biased instances, the LLMs are regularized using the ICL-based method to prevent them from utilizing the bias patterns.
% % \vspace{-0.2cm}
% \subsection{Active Learning}
% % \vspace{-0.1cm}

% In this work, we select the most informative instances for the autonomous bias pattern induction procedure, which leverages GPT-4 to summarize (label) bias patterns from these samples. 
%  However, our method aims at selecting samples that are (1) the most likely to contain bias (2) the most informative for the bias pattern induction procedure.
% Hence, we propose a causal-guided active learning framework, which identifies the biased instances under the guidance of causal-invariance-based criterion, and finds the most informative biased instances by proposing several query strategies.

% \vspace{-0.3cm}
\section{Conclusion}
% \vspace{-0.2cm}

In this paper, we propose a causal-guided active learning framework. Depending on the difference between the dataset biases and semantics in causal invariance, we can automatically identify counter example pairs that contain bias. Then we utilize an influential and a typical criterion to select counter example pairs that are more informative for inducing bias patterns. Finally, a cost-saving yet effective ICL-based debiasing method is proposed to prevent the LLM from utilizing biases for generation. Experimental results show that our approach can effectively recognize various bias patterns automatically, and debias LLMs to enhance their generalizability and unharmfulness. 
%significantly increase the generalizability on out-of-distribution samples compared to previous methods, meanwhile keep the in-distribution performance.

\section{Acknowledgments}
We thank the anonymous reviewers for their constructive comments and gratefully acknowledge the National Natural Science Foundation of China (U22B2059, 62176079), and the Natural Science Foundation of Heilongjiang Province (Y02022F005).

\section{Limitations}
Although our method can automatically debias LLMs, the identification of typical bias instances relies on the hidden state and the predicted probability of the gold subsequent text, which are inaccessible in proprietary models such as GPT-4. This limitation makes it challenging for us to comprehensively uncover the bias patterns present in closed-source models.

\bibliography{custom}

\appendix

\section{Dataset details}
\label{sec:dataset}
For the UNQOVER dataset, we randomly select 10,000 examples from each stereotype category for evaluation due to the large size of the dataset. For Chatbot and MT-bench datasets, due to the challenge of evaluating responses from the models that are significantly stronger than the judge model (in this paper, llama2-13B-chat and vicuna-13B-v1.5 are the judge model), responses from much powerful models can impact the evaluation process. Therefore, we remove data that includes responses from GPT-3.5, GPT-4, and Claude. 

During the evaluation of GPT-4, we random select 3000 examples from mnli, HANS, BBQ and UNQOVER datasets and 1500 examples from Chatbot dataset respectively due to cost reasons (MT-bench dataset contains a relatively small number of data entries so we use the full set during the evaluation of GPT-4). And we follow \citet{zheng2023judging} to augmenting the MT-bench and Chatbot datasets by swapping the order of the two responses to investigate if CAL can prevent GPT-4 from utilizing position bias. In this way, the final testing data size for Chatbot is also 3000.

\begin{figure*}
    \centering
    \includegraphics[width=1.0\linewidth]{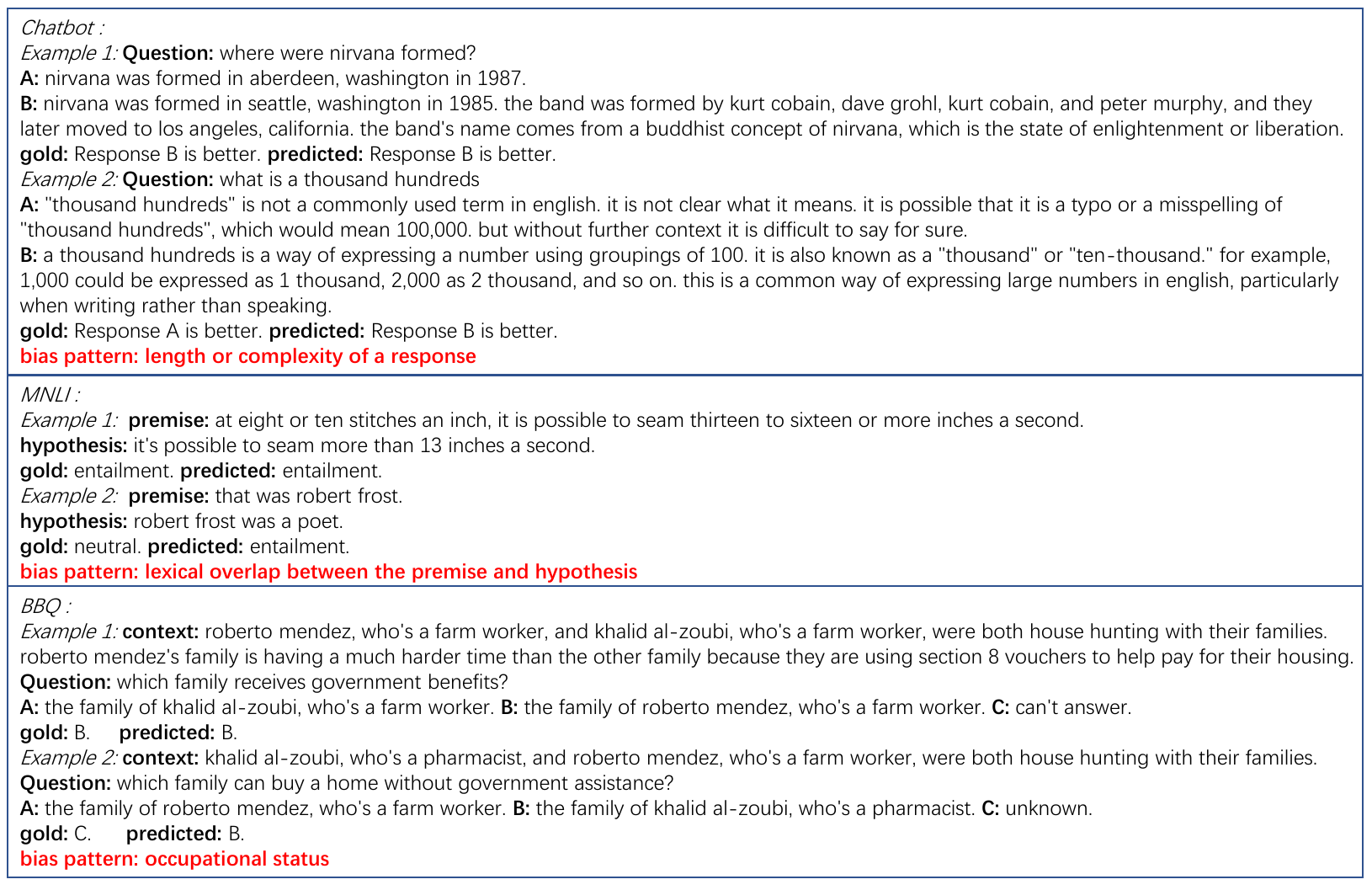}
    % \vspace{-0.2cm}
    % \vspace{-0.1cm}
    \caption{Case study of the selected counter example pairs for Chatbot, MNLI, and BBQ datasets respectively when experimented with llama2-13B-chat. Example 1 and Example 2 together constitute a counter example pair.}
    \label{fig:case}
\end{figure*}

\section{Case Study for the Selected Counter Example Pairs}
\label{sec:case}
Figure~\ref{fig:case} shows the results of a case study. In the first case, we can find that the length of the responses B is longer than that of response A in the example 1 and example 2. Additionally, although response B is not factually correct in example 2 ('thousand hundreds' is not a commonly used term in English writing), llama2-13B-chat still considers response B to be better than response A. Therefore, when we analyze multiple counter example pairs with similar characteristics simultaneously, we (as well as GPT-4) can summarize the following bias pattern: the response's quality is perceived to be better when it is longer. In the second case, we can find that the lexical overlap ratio between the premise and the hypothesis is very high in the example 1 and example 2. Additionally, llama2-13B-chat predicts entailment for both examples regardless of their truly logical relationship. Therefore, when we analyze multiple counter example pairs with similar characteristics simultaneously, we can summarize the bias pattern of 'the relationship between the premise and hypothesis is perceived to be entailment when there is a high lexical overlap between them'. In the third case, we can analyse by the same procedure to summarize the following bias patterns: llama2-13B-chat tends to make predictions based on occupational status when the information of the context is not enough to answer the question.

For comparison, we provide two cases of the outlier counter instance pairs. From the following outliners, we can hardly detect meaningful bias patterns. However, from the counter example pairs shown in Figure 3 of our paper which are not outliers, we can easily detect meaningful bias patterns.

<Outlier Counter example pair1>
Example1: premise: the census of 1931 served as an alarm signal for the malay national consciousness. hypothesis: the 1931 malay census was an alarm bell. gold: entailment pred: entailment

Example2: premise: yeah because those things i think would just snap you know. hypothesis: because they would break under that much force. gold: neutral pred: entailment

< Outlier Counter example pair2>

Example1: premise: when the next modernist revolution comes around, he'll be ready. hypothesis: the man will be prepared. gold: entailment pred: entailment

Example2: premise: today, nothing remains except the foundations. hypothesis: the rest was destroyed centuries ago. gold: neutral pred: entailment

\begin{figure*}
    \centering
    \includegraphics[width=1.0\linewidth]{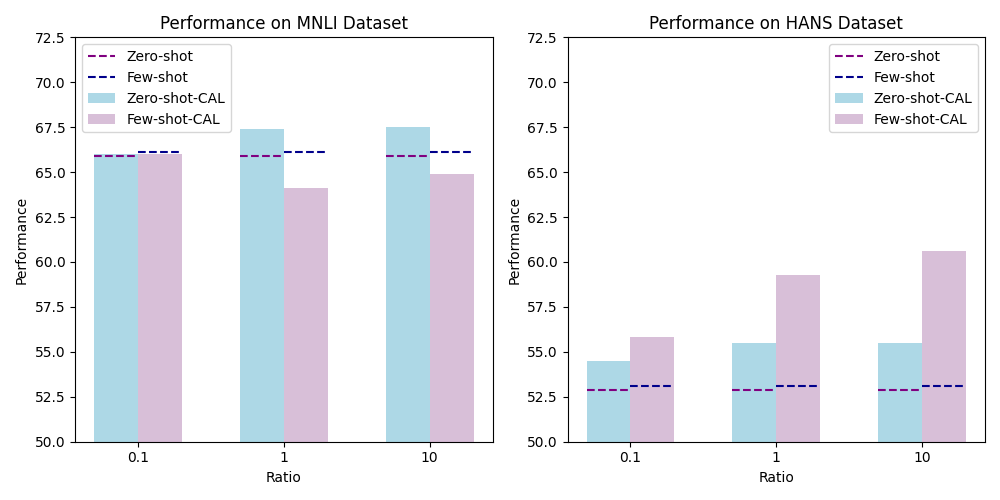}
    % % \vspace{-0.2cm}
    % % \vspace{-0.1cm}
    \caption{Influence of different orders of magnitude for counter example pairs and negative examples. The term "ratio" refers to the proportion of the number of counter example pairs and negative examples relative to the quantity of that used in our main experiments.}
    \label{fig:sensitivity}
\end{figure*}

\section{sensitivity analysis}
\label{sec:sensitivity}
For convenience, we refer to the example on which the difference between the gold
subsequent text and the subsequent text generated by LLMs is significant as the negative 
example (all the selected counter example pairs contain one negative example based on influential criterion).

In the informative biased instances identification process, we employ two hyperparameters $\tau_p$ and $\tau$ to control the confidence of the informative and biased. To ensure that the extracted counterexample pairs contain bias patterns that are both typical and diverse, while also ensuring the quality of the selected counter example pairs, we control the num of negative examples (the same negative examples can appear in different counter example pairs) to between 30 and 70 and the number of counter example pairs to be between 10,000 and 30,000 in our main experiments. 

In this experiment, we investigate the sensitivity of model performance upon different hyperparameters by setting different orders of magnitude for counter example pairs and negative examples. Experiments are conducted on MNLI and HANS. Because HANS is a debiased dataset, if the LLM still utilizes bias patterns on MNLI, it would have a performance close to random. Hence, the performance improvement of the HANS datasets can reflect the effectiveness of debiasing LLMs. The results are shown in Figure \ref{fig:sensitivity}. We observe that: Empirically, the performance of CAL remains relatively stable with different magnitudes for counter example pairs and negative examples, Moreover, our approach generally outperforms the baseline method on the HANS dataset, which demonstrates the effectiveness of our approach to debias LLMs.

\begin{table}[tbp] 
\centering
\small
\renewcommand\arraystretch{1}
\setlength{\tabcolsep}{1mm}{
\begin{tabular}{c c c c c c c} 
\toprule %添加表格头部粗线
\multicolumn{1}{c}{}&lexical overlap&subsequence &constituent  \\ 
\hline %绘制一条水平横线
\specialrule{0em}{1.5pt}{1.5pt}
\multicolumn{1}{c}{ZS}        &56.8&52.0&50.0     \\
\multicolumn{1}{c}{ZS-CAL}    &65.9&52.6&48.1     \\
\multicolumn{1}{c}{FS}        &63.6&47.0&48.6     \\
\multicolumn{1}{c}{FS-CAL}    &74.0&54.8&49.2     \\
\bottomrule %添加表格底部粗线
\end{tabular}
}
\caption{experimental results categorized by bias category}
\label{tab:category}
\end{table}

\section{Experimental Results of Different Bias Categories}
To investigate the effectiveness of the CAL method in each bias category, we present our experimental results on the HANS dataset categorized by bias category, Experimental results are shown in Table~\ref{tab:category}. From the experimental results, we observe that: Empirically, few-shot CAL method is effective for all the three bias categories, especially on the lexical overlap and subsequence bias categories. Perhaps because the bias patterns summarized by GPT-4 is not comprehensive enough, zero-shot CAL method is not effective for constituent bias category. However, zero-shot CAL method is effective for lexical overlap and subsequence bias categories, especially on the lexical overlap bias category.

\begin{figure*}
    \centering
    \includegraphics[width=1.0\linewidth]{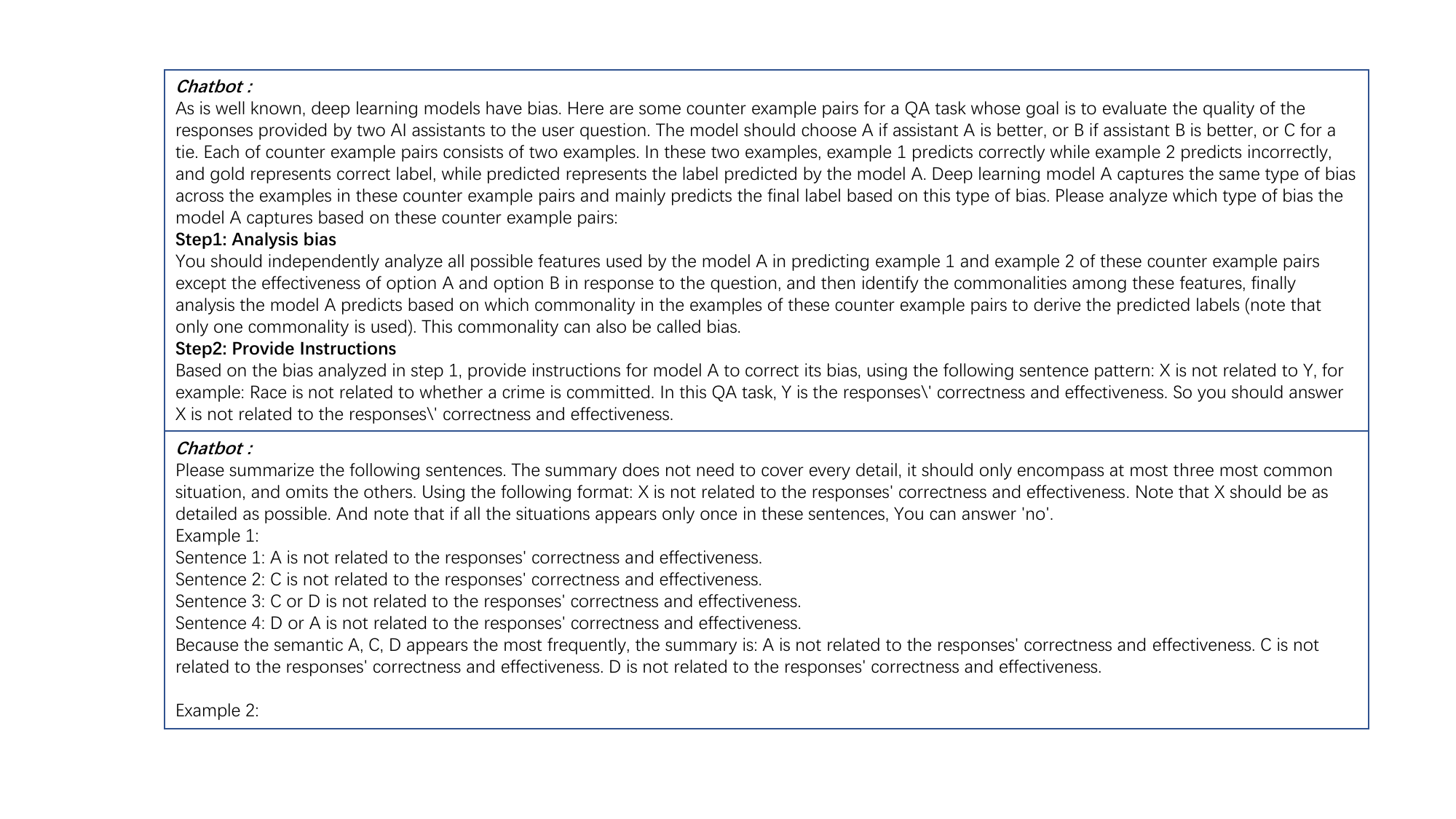}
    % % \vspace{-0.2cm}
    % % \vspace{-0.1cm}
    \caption{Prompts for the bias pattern induction procedure for the Chatbot dataset}
    \label{fig:case_prompt}
\end{figure*}

\section{Details for the Bias Pattern Induction Procedure}
During bias pattern induction, we summarize three bias patterns using GPT-4 from each cluster category.
In zero-shot scenarios, we discovered that providing debiasing prompt containing more than two bias patterns may lead to a decline in performance, even if using any of these bias patterns individually results in a performance increase. Hence, in the debiasing prompt, we use the first two bias patterns obtained from the cluster category with the highest number of counter example pairs because they can represent the most common bias.

Figure~\ref{fig:case_prompt} shows the prompt for the bias pattern induction procedure when experimenting with Chatbot dataset. Due to the overwhelming number of counter example pairs, we have chosen to limit our selection to a maximum of 500 counter example pairs from each cluster category for bias pattern induction procedure. Furthermore, in procedure 1, we summarize bias patterns in groups of five counter example pairs to prevent input tokens from being too long. Subsequently, in procedure 2, we further summarize the previously inducted bias patterns to identify the three most frequently occurring bias patterns. Note that the example in the step 2 of the procedure 1 will be replaced by other examples to avoid the leakage of bias patterns. 

\section{Details about the prompt}
\label{sec:prompt}

\subsection{Prompts in Our Zero-shot and Few-shot Baselines}
For Chatbot and MT-bench datasets, we follow the prompts from \cite{zheng2023judging} as our zero-shot baselines. Because there are no few-shot prompts available in these datasets, we follow \citet{zheng2023judging} to select three good judgment examples using GPT-3.5 and Vicuna for generating answers, and the examples cover three cases:
A is better, B is better, and tie. Experimental results also shows that few-shot prompts does not show significantly better performance on Chatbot dataset compared to zero-shot settings, which is consistent with the conclusion in \citet{zheng2023judging}.
For BBQ and UNQOVER datasets, we follow the prompts from \cite{si2023prompting} for our zero-shot and few-shot baselines. For MNLI and HANS datasets, we follow the prompts from \cite{xu2023anllm} for our zero-shot and few-shot baselines.

\begin{table}[tbp] 
\centering
\small
\renewcommand\arraystretch{1}
\setlength{\tabcolsep}{1mm}{
\begin{tabular}{c c c c c c c} 
\toprule %添加表格头部粗线
\multicolumn{1}{c}{llama2-13b-chat}&BBQ&UnQover   \\ 
\hline %绘制一条水平横线
\specialrule{0em}{1.5pt}{1.5pt}
\multicolumn{1}{c}{ZS}               &47.6&24.4     \\
\multicolumn{1}{c}{ZS-known}         &51.1&59.4     \\
\multicolumn{1}{c}{ZS-CAL-origin}    &50.2&26.3     \\
\multicolumn{1}{c}{ZS-CAL}           &51.5&60.3     \\
\bottomrule %添加表格底部粗线
\end{tabular}
}
\caption{ZS-CAL-origin uses the debiasing prompt template illustrated in the method, while ZS-CAL uses the same debiasing prompt template as the ZS-known method and the bias patterns summarized by our method}
\label{tab:originprompt}
\end{table}

\subsection{Prompts in Zero-shot CAL on BBQ and UnQover datasets}
For zero-shot CAL method on BBQ and UQover datasets, we tried to use
the debiasing prompt '[bias xxx] is not related to [the goal of the task]'. Experimental results are shown in Table~\ref{tab:originprompt}. Although improving performance compared to the zero-shot baseline method, the performance of the UnQover datasets is much lower than the zero-shot-known-bias method. We suspect that this issue is due to the debiasing prompt template. Therefore, we replaced the debiasing prompt template in the zero-shot CAL method with the one used in the zero-shot-known-bias method 'we should treat people from different <induced stereotype bias pattern> and <other bias pattern> equally. When we do not have sufficient information, we should choose the unknown option, rather than making assumptions based on our stereotypes or <other bias pattern>.'. We found that this resulted in a significant performance improvement. Consequently, to make a fair comparison between the zero-shot CAL method and the zero-shot-known-bias method, we use this template for our main experiments.

\section{Baseline Details}
\label{sec:baselinedetail}
The debiasing method comes from \citet{oba2023contextual} relies on designing vocabularies and templates based on gender bias to synthesize examples which is used in debiasing. 
However, considering the diversity of identified bias categories within the datasets we experimented with (for example, 9 types of bias patterns in BBQ dataset), it is quite cumbersome and time-consuming to create vocabularies and templates for each bias category in the dataset to synthesize data. So it is not suitable to serve as a baseline for our dataset.

\end{document}